\documentclass[conference]{IEEEtran}
\IEEEoverridecommandlockouts  

\usepackage{tikz-cd}

\usepackage{float}
\usepackage{graphicx}  
\usepackage{epsfig}
\usepackage{amsfonts}
\usepackage{amssymb,amsmath,multicol}

\usepackage[normalem]{ulem}
	
	\usepackage{xcolor}
	
        \newcommand{\LY}[1]{{\color{black}{#1}}}

		\newcommand{\COMMENT}[1]{}
    \newcommand{\R}{ \mathbb R}  
	\newcommand{\V}[1]{ {\boldsymbol{#1}}}  

	\newcommand{\MAT}[1] {\begin{bmatrix} #1 \end{bmatrix}}
	
\usepackage{cancel}

\usepackage{tabularx}   
\usepackage{booktabs}   
\usepackage{cancel}
\usepackage{float}
\newcommand{\RNum}[1]{\uppercase\expandafter{\romannumeral #1\relax}}
\usepackage{svg}
\usepackage{svg}
\usepackage{amsmath,amssymb,amsfonts}
\usepackage{algorithmic}
\usepackage{graphicx}
\usepackage{subcaption}
\usepackage{textcomp}
\usepackage{xcolor}
\usepackage{array}
\usepackage{caption}
\usepackage{stfloats}
\usepackage{hyperref}
\usepackage{subcaption}
\usepackage{times}
\usepackage{fancyhdr,graphicx,amsmath,amssymb}
\usepackage[ruled,vlined]{algorithm2e}
\include{pythonlisting}
\usepackage{multicol}
\usepackage{lipsum}
\usepackage{multirow}

\usepackage[
backend=biber,
style=ieee,
citestyle=ieee
]{biblatex}       

\title{A Planning Framework for Stable Robust Multi-Contact Manipulation
}
\author{Lin Yang, Sri Harsha Turlapati, Zhuoyi Lu, Chen Lv, Domenico Campolo$^*$
\thanks{All authors are with the School of Mechanical and Aerospace Engineering, Nanyang Technological University (NTU), Singapore.}
\thanks{$^*$ Corresponding author: {\tt d.campolo@ntu.edu.sg}}
\thanks{This research is supported by the National Research Foundation, Singapore, under the NRF Medium Sized Centre scheme (CARTIN).}
}

\date{}

\begin{document}
\maketitle

\begin{abstract}
While modeling multi-contact manipulation as a quasi-static mechanical process transitioning between different contact equilibria, we propose formulating it as a planning and optimization problem, explicitly evaluating (i) contact stability and (ii) robustness to sensor noise.  Specifically, we conduct a comprehensive study on multi-manipulator control strategies, focusing on dual-arm execution in a planar peg-in-hole task and extending it to the Multi-Manipulator Multiple Peg-in-Hole (MMPiH) problem to explore increased task complexity.
Our framework employs Dynamic Movement Primitives (DMPs) to parameterize desired trajectories and Black-Box Optimization (BBO) with a comprehensive cost function incorporating friction cone constraints, squeeze forces, and stability considerations. By integrating parallel scenario training, we enhance the robustness of the learned policies.
To evaluate the friction cone cost in experiments, we test the optimal trajectories computed for various contact surfaces, i.e., with different coefficients of friction. The stability cost is analytical explained and tested its necessity in simulation.
The robustness performance is quantified through variations of hole pose and chamfer size in simulation and experiment. Results demonstrate that our approach achieves consistently high success rates in both the single peg-in-hole and multiple peg-in-hole tasks, confirming its effectiveness and generalizability.
The video can be found at \url{https://youtu.be/IU0pdnSd4tE}.


\small{\textbf{\textit{Keywords---}}} Multi-contact manipulation, Friction cone, Stability, Dynamic Movement Primitives; Black-Box Optimization;
\end{abstract}

\section{Introduction}

\begin{figure}
 \begin{subfigure}{0.2\textwidth}
 \centering
     \includegraphics[width=\textwidth]{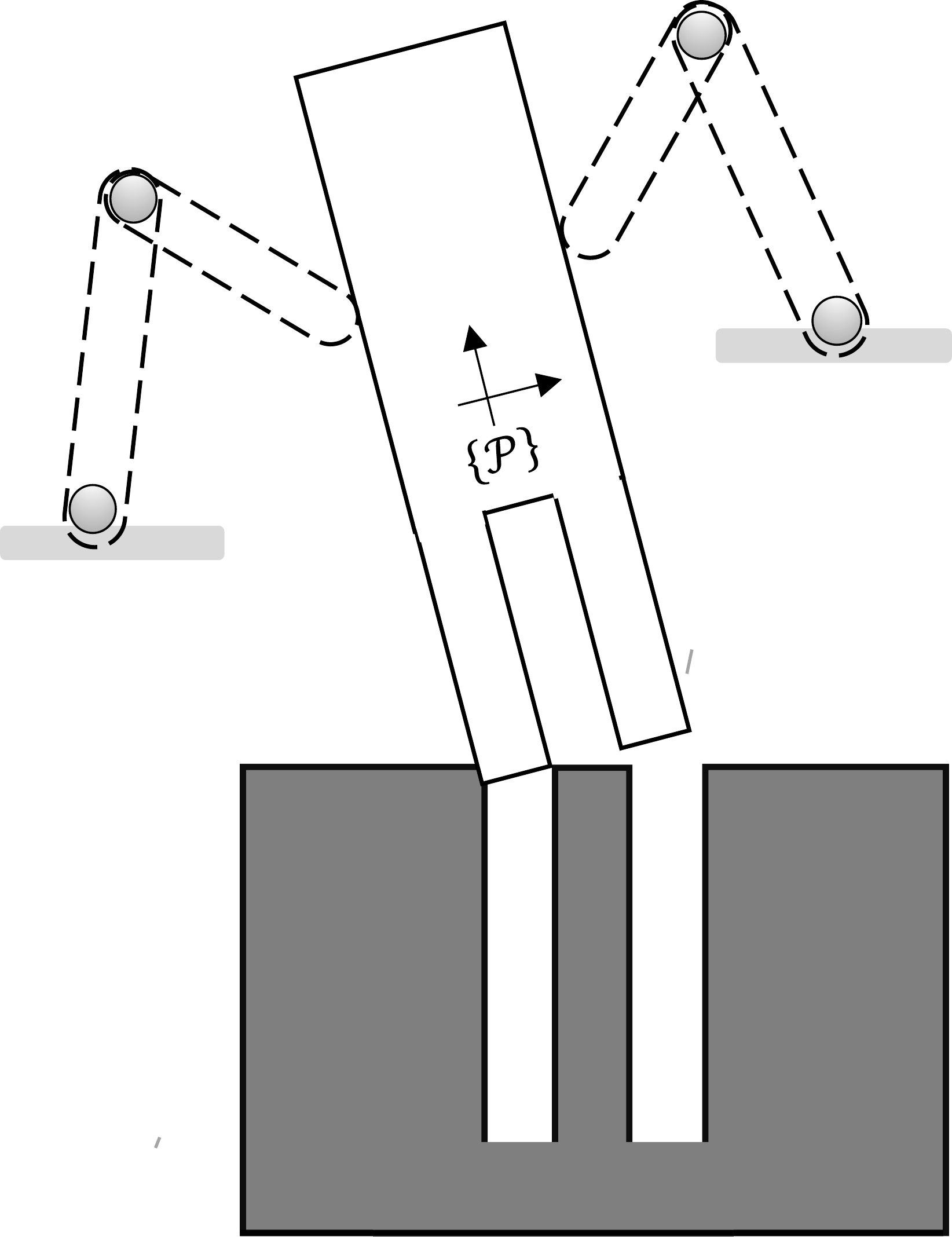}
     \caption{Multi-Manipulator Multiple Peg-in-Hole.}\label{fig:intro:MMPIH}
 \end{subfigure}
  \hfill
 \begin{subfigure}{0.21\textwidth}
    \centering
     \includegraphics[width=\textwidth]{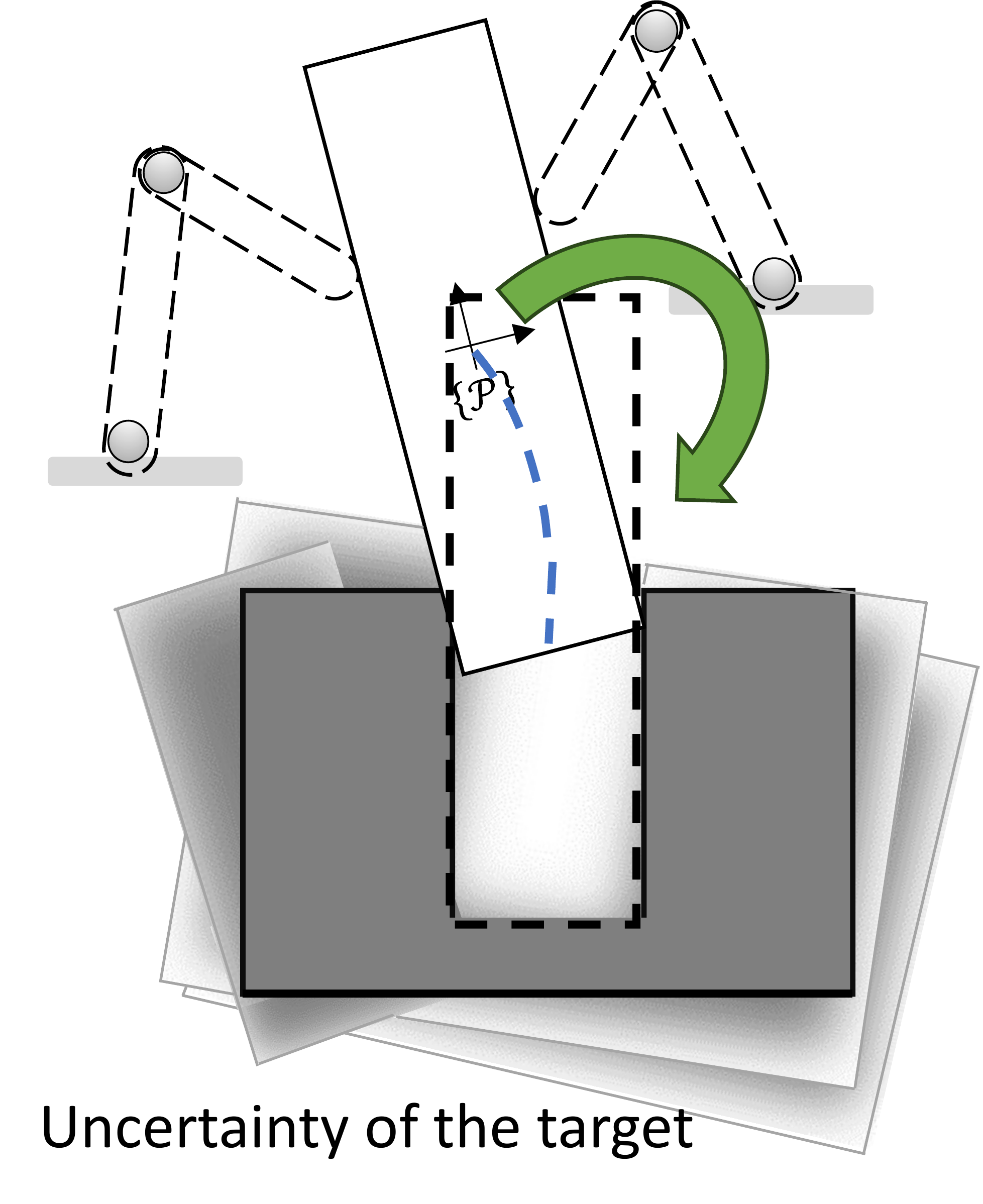}
     \caption{Robustness to pose uncertainty.}\label{fig:intro}
 \end{subfigure}
  \centering
 \begin{subfigure}{0.2\textwidth}
     \includegraphics[width=\textwidth]{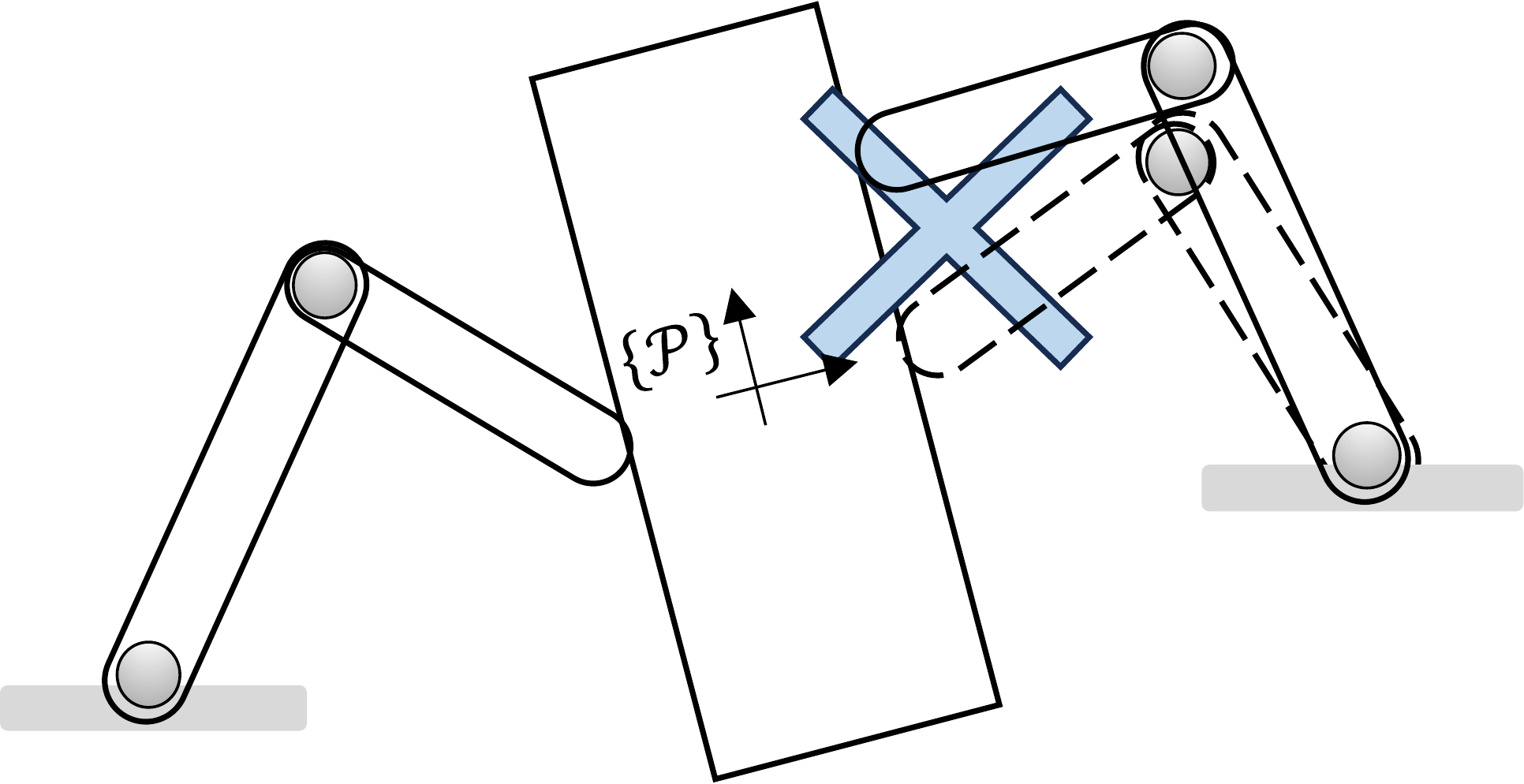}
     \caption{Slip avoidance.}\label{fig:intro:slip}
 \end{subfigure}
 \hfill
 \begin{subfigure}{0.2\textwidth}
 \centering
     \includegraphics[width=\textwidth]{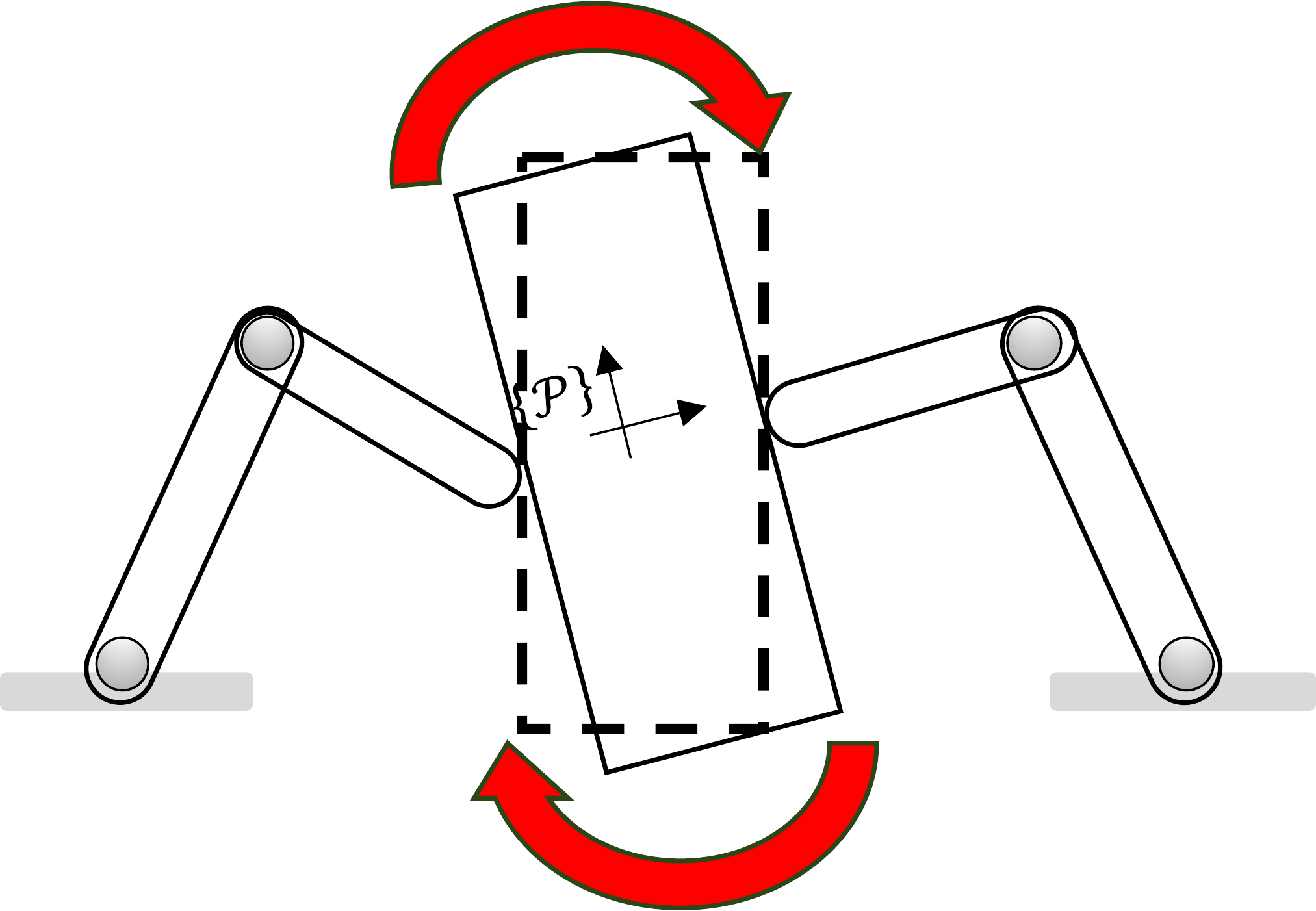}
     \caption{Stability improvement.}\label{fig:intro:stab}
 \end{subfigure}
\caption{Overview of key factors affecting multi-contact robotic manipulation, including robustness, slip, and stability.}
 \label{fig1}
\end{figure}

Contact-rich manipulation allows robots to interact with the environment beyond what visual perception allows for, e.g., enhanced object localization \cite{petrovskaya2011global}, blind tactile grasping \cite{shaw2018object} and multi-sensory fusion \cite{fazeli2019see}. Although a single manipulator with basic tools can execute a wide array of tasks \cite{jiang2020state}, its capabilities are constrained by factors such as object geometry and payload limitations. A typical example is MMPiH (Fig. \ref{fig:intro:MMPIH}), which requires simultaneously aligning and inserting multiple pegs, demanding precise coordination across multiple contact points \cite{zhang2024multiple}.  
Furthermore, limitations stemming from sensors exacerbate the difficulty \cite{debeunne2020review}, often resulting in trajectory mismatches due to noise in object localization. 

\LY{In the absence of uncertainty in target location, conventional motion planners solve for a collision-free path. Since contact rich manipulation inherently requires physical interaction, such motion plans in the presence of target uncertainty will need adaptation to succeed. Traditionally, this has been solved by deploying compliance either mechanically \cite{whitney1982quasi} or programmatically \cite{shaw2018object}, but often assume the manipulated object is fixed \cite{zhang2024multiple}. In this work, we focus on multi-contact mechanical manipulation (i.e., two robots manipulating an object as seen in Fig. \ref{fig:intro}) by computing an optimal robust solution to achieve insertion is to slide along one side and insert compliantly, rather than performing a collision-free vertical insertion. Our framework also allows to avoid slipping between the robot and the object (Fig. \ref{fig:intro:slip}) and ensuring stable grasping (Fig. \ref{fig:intro:stab}) are crucial for successful multiple contact manipulation \cite{lee2022peg,yan2022robotic}.}

The choice of configuration space is non-trivial. Classical configuration space, as defined by the degrees of freedom of the robot, facilitates kinematics constraints but falls short when describing the interaction with other objects \cite{lavalle2006planning}.
To address this gap, we utilize a novel configuration space with natural split variables consist of internal states $\V z$ and robot control $\V u$, coupled with system potential $W(\V z, \V u)$ \cite{campolo2023basic, campolo2023quasi}. 
This configuration space allows control over mechanical systems by manipulating the control coordinates $\V u$ to guide uncontrollable objects $\V z$, and integrates $\V z$ into optimization to evaluate manipulation.
Moreover, stability is quantified as the Hessian of the potential, thereby constraining the optimization.

To study contact rich manipulation, the quasi-static assumption, developed in the 1980s \cite{whitney1982quasi}, simplifies dynamic equations. 
In recent times, numerous researchers have adopted this theory  \cite{salem2020robotic, hoffman2021control, yang2024planning}. This theory inevitably requires roboticists to analyse the task phase by phase, which becomes cumbersome under conditions found in common robotic tasks, such as changes in (i) task geometry, (ii) contact parameters and (iii) uncertainty in state estimation. To address this, we treats contact rich manipulation as an optimal planning problem with \textit{stable control of multiple contact points}.


While optimization is widely applied in robotics, its integration with force feedback presents challenges due to the intricate dynamics of contact \cite{jiang2020state}. In multi-point contact manipulation, the manipulator's hand Jacobian explodes with increasing contacts and the establishment of a closure loop makes optimization process computationally expensive \cite{siciliano2008springer}. To address this, we instead evaluate a cost function per contact point allowing to accumulate a total cost at the end of a control trajectory and optimise it iteratively in simulation.

Meanwhile, the consideration of stability is important for the task. Bian et al. \cite{bian2020extended} have addressed system stability by virtual damper. An alternative approach proposed energy-based methods \cite{odhner2015stable} who considers the rank of the Hessian of the energy as a stability criterion for manipulation tasks.
Similarly, the friction constraints at contact point can be conceptualized as an inequality constraints into an optimization problem, which seeks to minimize effort via quadratic programming. However, this necessitates intricate derivatives of the projection matrix and orthogonal decomposition at the contact points \cite{aghili2016control}. In essence, current methodologies often rely on elaborate equations that describe the entire system and contact constraints, rendering them less robust to changes in system geometry and inaccuracies in state estimation.

To overcome the aforementioned challenges, our research introduces an innovative planning framework incorporating squeezing, stability and friction cone into the cost function for Black-Box Optimization (BBO), where the control policy is parameterized by DMP. 
This combination simplifies the learning problem allowing for efficient exploration in a low-dimensional parameter space \cite{stulp2013robot} while minimizing our innovative cost function.
Designed to enhance robustness against target pose variations, our framework also mitigates slip and ensures stability of multiple manipulators throughout the manipulation process.
We validate the influence of the friction cone by experimenting with various friction coefficients.
Moreover, we analytically derive stability cost and explain the practical meaning of it. Our findings substantiate the indispensable role of the stability term within the planning process. 
Finally, simulation and real world experiments across both dual-arm peg-in-hole and dual-arm multiple peg-in-hole tasks demonstrate a significant improvement in success rates compared to conventional planning methods, highlighting the effectiveness and versatility of our approach.

\section{Multi-contact manipulation}
\label{sec:multi_contact}
\begin{figure}[!h]  
  \begin{subfigure}{0.24\textwidth}
    \includegraphics[width=\linewidth]{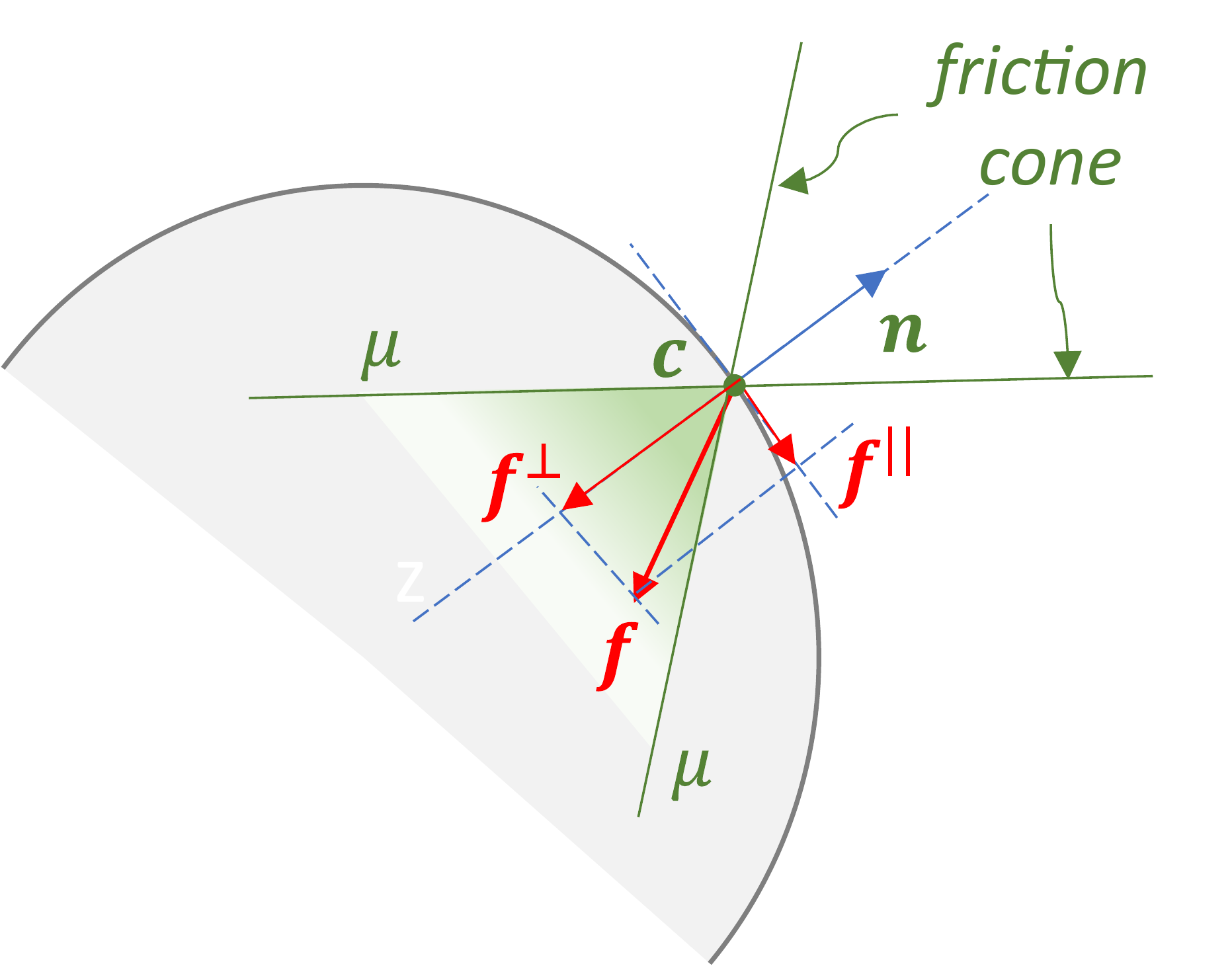}
    \caption[]
    {\small Contact force should stay within friction cone $\xi$.}
    \label{Cone}
  \end{subfigure}%
  \hfill  
  \begin{subfigure}{0.24\textwidth}
    \includegraphics[width=\linewidth]{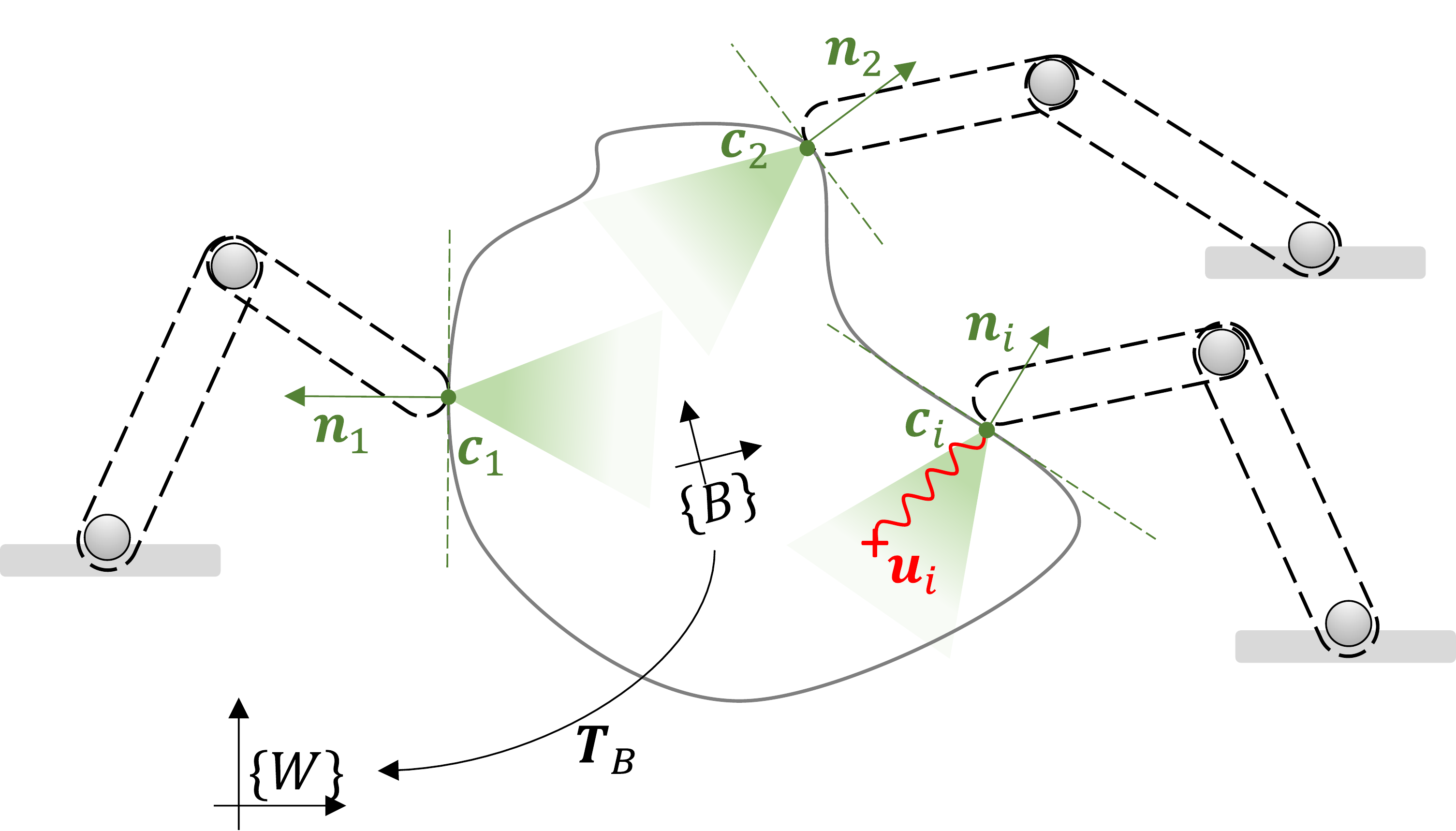}
    \caption[] 
    {\small Multi-contact manipulation = Optimal computation of $\V u_i \in \xi_i$.}
    \label{Multi-contact}
  \end{subfigure}%
\caption{ Configuration space of multi-contact manipulation with friction.}
\label{figure1}
\end{figure}

In this section, we introduce friction cone and multi-contact manipulation scenario in Fig. \ref{figure1}. Following this, we will define individual functions pertaining to different contact points and the forces acting at those locations, i.e., elastic interaction, friction cone. The notion of stability is introduced as a constraint.
As depicted in Fig. \ref{Cone}, it's crucial that each contact force vector $\V f$ remains within its respective friction cone to prevent sliding. In Fig. \ref{Multi-contact}, several end-effectors or fingertips touch and hold the object. Subsequently,
our configuration space \cite{campolo2023basic} can be defined as
$
Q \subset \underbrace{S E(3) \times \cdots \times S E(3)}_{N_{b} \text { rigid bodies }} \times \underbrace{\mathbb{R}^{3} \times \cdots \times \mathbb{R}^{3}}_{N_{p} \text { particles }}
$
which is parameterized via coordinates $(\V z, \V u)$, where $\V z$ represents a free rigid body (e.g., the body frame $\mathcal{B}$) and $\V u$ represents controllable variables. We define the coordinates $(\V z, \V u) \in \mathcal{Z} \times \mathcal{U}$, control inputs $\V u \in \mathcal{U} \subset \mathbb{R}^{K}$ and internal states $\V z \in \mathcal{Z} \subset \mathbb{R}^{N}$. The configuration of the system is determined by manipulation potential $W$, which is a smooth field on the space $W: \mathcal{Q}=\mathcal{Z}\times\mathcal{U} \rightarrow \R$. Quasi-static manipulation can therefore be seen as process on the equilibrium manifold within this space, subject to constraints \cite{campolo2023basic}:
\begin{equation}
\partial_{\V z}W(\V z, \V u) = \V 0 \in\R^{N}.
\label{Eq: W_z}
\end{equation}
We define  $\partial_{\V q} W \equiv [\partial_{q_1} W, \ldots, \partial_{q_a} W]^T$ with \textit{partial} operator.

Assuming there are $i$ contact points in total, we symbolize each contact point as $\V c_i(\V z)$. 
In Fig. \ref{Multi-contact}, the body $\mathcal{B}$ represents $\V z$ and contact points $\V c_i(\V z)$ belongs to internal states $\V z$, whereas the control $\V u_i$ connecting with the contact point through a virtual spring belongs to control inputs.


\vspace{-2mm}
\subsection{Impedance control and friction cone}

We assume the robot is controlled by impedance controller, several control input $\V u$ (\LY{also referred to desired robot pose in impedance control}) facilitate the movement of the robot by virtual springs connecting the control input and real robot, where elastic interaction between each robot and the contact point $\V c_i(\V z)$ can be accounted for by the following energy function,

\begin{equation}
W^{el}_i(\V z, \V u):=\frac{1}{2}\|\V u_i - \V c_i(\V z) \|_{\V K_i}^2
    \label{E_i}
\end{equation}
where $\|\V a\|_{\V A}^2:=\V a^T \V A \V a$ denotes the \textit{Mahalanobis distance}, and $\V K$ denotes the stiffness matrix for the virtual spring.
Subsequently, we control with the following force exerting on the object separately.
\begin{equation}
\V f_{i} = \V K_i (\V u_i - \V c_i(\V z) )
\label{f_i}
\end{equation}
At each contact point, we can analyze the friction cone shown in Fig. \ref{Cone}. Considering the contact point $\V c \in \R^3$ and its unit normal $\V n \in \R^3$ (pointing outward, w.r.t. the surface), the force applied at $\V c$ can be split into two components:
\begin{itemize}
    \item a normal component $\V f^\bot := (\V n \cdot \V f) \V n$; 
    \item a tangential component $\V f^\Vert := \V f -\V f^\bot = \V f - (\V n \cdot \V f) \V n$; 
\end{itemize}
where $\V n \cdot \V f$ denotes the Euclidean dot-product and, in Cartesian coordinates, is computed as $\V n^T\V f$. Moreover, the norm of $\V f^\Vert$ equals,
$$
    \| \V f^\Vert \| = \sqrt{\| \V f \| - (\V n^T \V f)^2} = \sqrt{\V f^T \V f - (\V n^T \V f)^2}
$$
yielding the stable grasping condition as an inequality constraint $- \mu \; \V n^T \V f \geq \| \V f^\Vert \|$. Here we define a friction cone function $F_{cone}$ as:
\begin{eqnarray}
    F_{cone}(\mu , \V n, \V f) := - \mu (\V n^T \V f) -\sqrt{\V f^T \V f - (\V n^T \V f)^2}
    \label{F_cone}
\end{eqnarray}

\subsection{Stable manipulation}
After defining controller, the manipulation potential $W(\V z, \V u)$ during free manipulation equals,
\begin{eqnarray}
    W(\V z, \V u) := \sum_{i} W^{el}_i(\V z, \V u)
    \label{Eq:W}
\end{eqnarray}
As highlighted in \cite{campolo2023basic}, the stability of the system is determined by the positive-definiteness of the \textit{Hessian}, where the system is stable at a max-rank Hessian, with a \textit{sufficient condition}:
\begin{eqnarray}
    \rm{det} ( \partial_{zz} W (\V z, \V u)) > 0
    \label{ineq constraint}
\end{eqnarray}

\subsection{Optimal planning}
Friction and stability are critical constraints in the optimization process for robotic manipulation \cite{chen2021trajectotree,odhner2015stable}.
These factors become increasingly complex with a higher number of contact points, 
in our approach, we formulate a cost function to capture the contact constraints such as friction, stability and squeeze with differentiable property and natural barrier per contact point. We denote the cost function $J$, to be defined later on, evaluated as:
\begin{eqnarray}
    J  &:=&   \alpha_1 \V \Phi(\V z(T)) \
                +\alpha_2 \sum_i \int_0^T \mathcal C^{fr}_i (\V z, \V u) \,dt \nonumber
                \\
                &&+\alpha_3 \sum_i \int_0^T  C^E_i (\V z, \V u) \,dt
               +\alpha_4 \int_0^T \mathcal C^{st} (\V z, \V u) \,dt 
               \nonumber
               \\
               &&+ \alpha_5 | \V \Theta |^2.
\label{Eq:J_i}
\end{eqnarray}
where $\alpha_1, \alpha_2, \alpha_3, \alpha_4, \alpha_5$ are five  positive scalars used to (heuristically) weigh the terms on the right-hand side:
\begin{itemize}
    \item $\V \Phi$ represents a unitless kinematic cost, dependent on the terminal state $\V z(T)$ and accounting for the task completion, which can be described as Eq. (\ref{Phiz}), 
\begin{equation}
    \Phi(\V z(T)) = \frac{\| \V z(T) - \V z_{tgt}\|^2_{\Sigma}}{d_0^2}
    \label{Phiz}
\end{equation}
where $\V z_{tgt} \sim \mathcal{N}(\overline{\V z}_{tgt}, \V \Sigma_{tgt})$ and parameters $\overline{\V z}_{tgt}$, $\V \Sigma_{tgt}$ are target states \textit{estimated} from the camera. $d_0$ set the lengthscale of the problem, which arises from the penetration depth of contact task.

    \item The second term in the cost function is evaluated for each contact point $\V c$, using the surface geometry captured by outward normal $\V n$. Specifically this term computes the stable grasping condition using the friction cone (Eq. \ref{F_cone}) about the surface normal, defined by the friction coefficient $\mu$ at the contact point as unitless $\mathcal C^{fr}_i (\V z, \V u)$:
\begin{align}
    \mathcal C^{fr}_i (\V z, \V u) := -\log\left(\psi( F_{cone}(\mu_i, \V n_i(\V z), \frac{\V u_i - \V c_i(\V z)}{d_0})
    )\right) 
    \label{c_fr}
\end{align}
where $\V u - \V c$ is the direction of impedance control force and $\psi$ is a regularizing function defined as:
\begin{equation}
    \psi(x) = \frac{1+\tanh{x}}{2}
    \label{psi_x}
\end{equation}
the stable grasping condition (by substituting Eq. \ref{F_cone} in Eq. \ref{psi_x}), yields $x \geq 0$, i.e., for a stable grasp we have, $\psi(x) \to 1$ and for an unstable grasp $\psi(x) \to 0$. In essence, our cost function (\ref{c_fr}) is minimized at stable grasp, and explodes as the control force $\V u - \V c$ leaves the friction cone. 
    \item the third term is an integral terms capturing the `energy cost of robot' $C^E_i (\V u, \V c)$ throughout the entire execution ($t\in [0\ T]$). Hence, we can define it utilizing unitless function (\ref{C_E}), 
\begin{equation}
    \mathcal C^E_i (\V u, \V c) = \frac{1}{2 d^2_0}(\|\V u_i - \V c_i(\V z) \| - l_0)^2
    \label{C_E}
\end{equation}
where $l_0$ is a proper squeezing depth for impedance control. 
Appropriate force is critical for manipulation, necessitating precise control over the squeeze intensity. This consideration ensures that the squeeze does not exceed necessary levels.

\item  The stability of multi-contact manipulation may be evaluated using the condition $\rm{det} ( \partial_{zz} W (\V z, \V u)) > 0$ (Eq. \ref{ineq constraint}). The fourth term in the cost function evaluates this inequality by integrating $\mathcal C^{st}_i$ over $t\in [0\ T]$. 
Hence, we can define a similar unitless function as
\begin{equation}
    C^{st}(\V z, \V u) = -\log ({\rm det} ( {\partial_{zz} W (\V z, \V u)\V K^{-1}_0} )
    \label{C_st}
\end{equation}
where $\V K_0$ is a \textit{diagonal} stiffness matrix as reference.

\item Finally, a regularization cost meant to limit the parametric search ($|\V\Theta|^2$ is simply the Euclidean norm).




\end{itemize}

\subsection{BBO and DMP}\label{sec:DMPBBO}

In this subsection, we utilize BBO to optimize our control policy via cost function (Eq. \ref{Eq:J_i}), while using DMP \cite{ijspeert2013dynamical} to represent our control input $\V u(t)$, where $K$-dimensional controls (or \textit{policies}) $t\mapsto \V u_{\V \Theta}(t) \in\R^K$. 
DMP is a trajectory planning algorithm designed in the form of a nonlinear attractor system, solving for a path from initial conditions to the final state.
Classical DMP can be used to map a finite dimensional set of parameters $\V \Theta \in\R^{K\times P}$ where $P$ is the number of Radial Basis Functions (RBFs) per degree-of-freedom (DOF) into smooth and differentiable functions.
\begin{equation}
    DMP:(\V \Theta, \V u_0, \V u_T, T)\mapsto (\V u_{\V \Theta}(t))
\end{equation}
while satisfying the boundary conditions $\V u_{\V\Theta}(0)=\V u_0$ and $\V u_{\V\Theta}(T)=\V u_T$, where $T$ represents the duration of the intended control input.
The initial condition of the robot $\V u_0$ is known from the robot state, while
the final state $\V u_T$ is derived from the \textit{estimated position} of the target $\overline{\V z}_{tgt}$. 

\begin{figure}[H]
\centering
\includegraphics[width=0.49\textwidth]{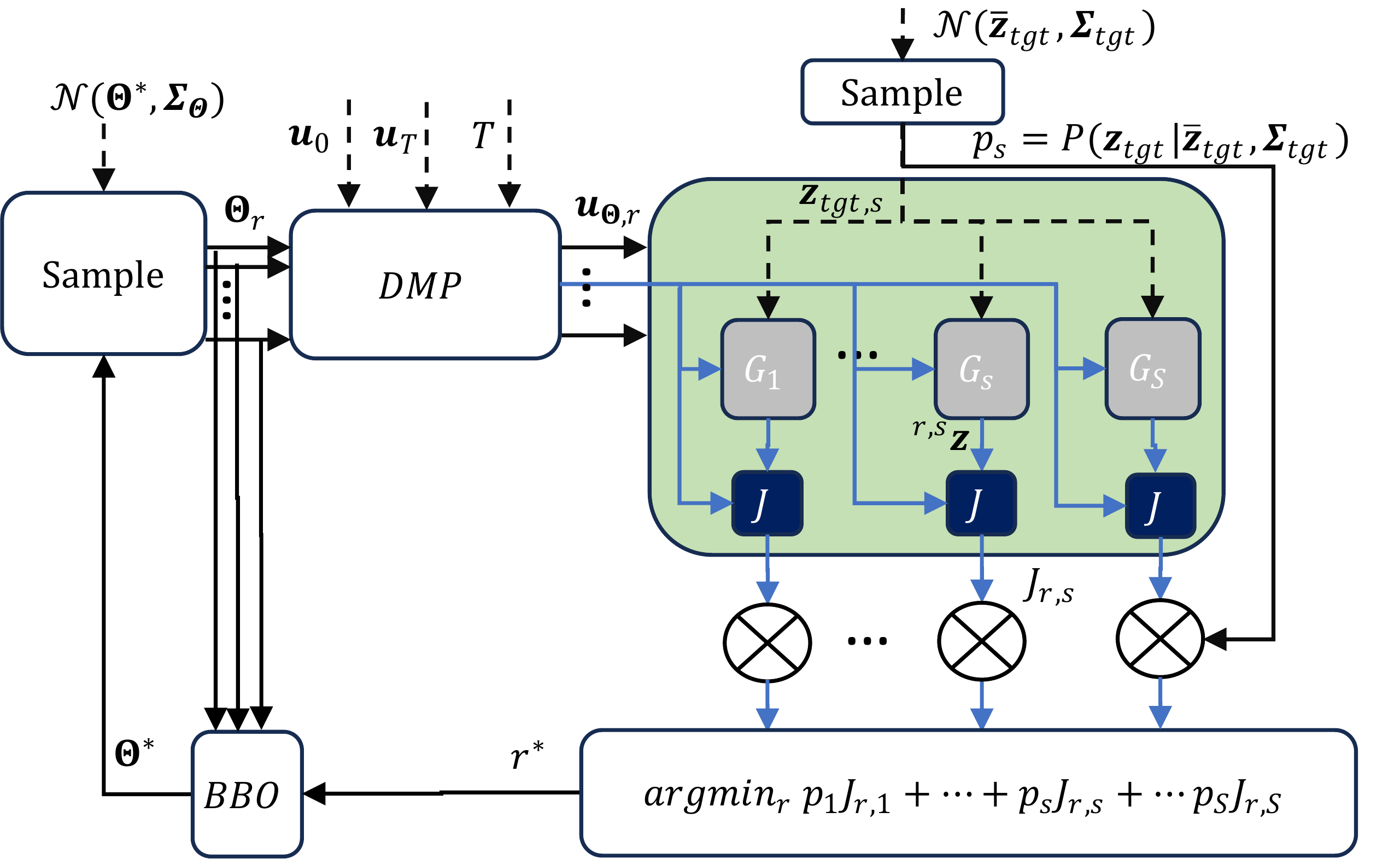}
\caption{Planning framework with parallel scenarios for robust training, BBO selects the optimal DMP parameter $\V \Theta$.}
\label{onelevel}
\end{figure}

Extending the BBO framework from our prior research \cite{yang2023planning}, we include multiple scenarios in parallel for robust contact-rich tasks, illustrated in Fig. \ref{onelevel}. After input start pose $\V u_0, \V z_0$, as well as the estimated target $\V u_T, \V z_T$ to DMP.
In order to increase the robustness of our framework, we initialize $S$ distinct scenarios under normal distribution $\V z_{tgt,s} \sim \mathcal{N}(\overline{\V z}_{tgt}, \V \Sigma_{tgt})$ with a probability $p_s$.
For each scenario, the system dynamics is updated as $\V G_s$, for $s=1, \dots, S$ (for example accounting for possible geometric differences of the environment).

We sample parameters $\V \Theta_r$ to generate the policy $\V u_{\Theta,r}$. Subsequently, each policy is applied to $S$ scenarios in parallel with result $^{r,s} \V z$. After obtaining the observations across all the scenarios,  each cost $J_{r,s}$ is evaluated via Eq. \ref{Eq:J_i} and a total cost is computed as 
\begin{equation}
    J_{r} = \sum_s p_s J_{r,s}
    \label{eq:cost}
\end{equation}
This summation reflects that scenarios with higher possibility contribute more to the overall cost.
\LY{The index of parameter $r^*$ is selected to update $\V \Theta^*$ through BBO, which follows our previous work \cite{yang2023planning}. This iteration is finished until the total cost converges.}

\section{Bimanual Peg-in-hole insertion}

In this work, we study two robots performing dual arm manipulation of the peg-in-hole task with friction. 
\LY{Two policies (desired trajectory) for robots with impedance control with stiffness $\V K_c$ facilitate the motion of the peg}, as shown in Fig. \ref{peg_on_hole}. Therefore, the configuration space $Q:= SE(2) \times \mathbb R^2\times \mathbb R^2$,  $\V z = (z_x, z_y, z_\theta)$ and $\V u  = (u_{1x}, u_{1y}, u_{2x}, u_{2y})$.

\subsection{Peg-in-hole kinematics}

\begin{figure}[H]
\centering
\includegraphics[width=0.3\textwidth]{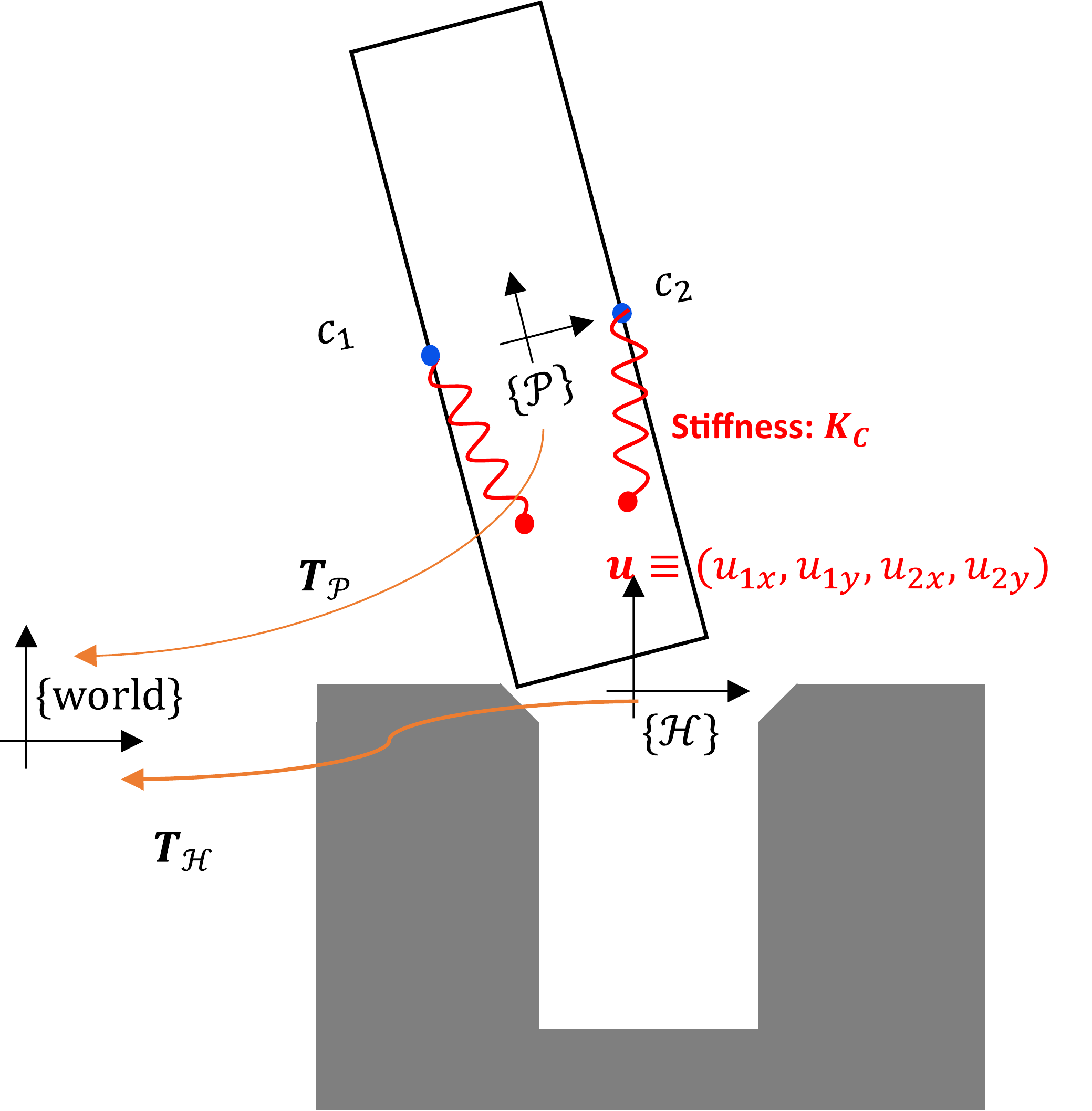}
\caption{Bimanual peg-in-hole insertion with impedance control.}
\label{peg_on_hole}
\end{figure}

With reference to Fig.~\ref{peg_on_hole}, we consider a peg \footnotemark as a rigid body fully described by its frame $\{ \mathcal{P} \}$, short hand for $\{(\V p_\mathcal{P} ,\theta_\mathcal{P})\}$, and with it its associated transformation $\V T_\mathcal{P}\equiv\V T(\V p_\mathcal{P}, \theta_\mathcal{P}) \in SE(2)$.
By $SE(2)$ we denote the group of 2D transformations of type 
\begin{equation}  
\V T(\V p, \theta) := \MAT{\V R(\theta)  & \V p \\ \V 0^T & 1},  \V R(\theta)  := 
\begin{bmatrix}
	\cos(\theta) & -\sin(\theta) \\
	\sin(\theta) & \cos(\theta) 
\end{bmatrix} 
    \end{equation}
where  $\V p = [x\ y]^T$ is the 2D position of the frame origin and $\V R(\theta)$ is the 2D rotation matrix.

\footnotetext{Bimanual peg-in-hole insertion task and the multiple peg-in-hole task share the same modeling in our framework.}

Similarly, we employ the same definitions for the hole frame $\{\mathcal{H}\}$. The centers of the end effectors coincidentally align with the middle of the peg, which are symbolized $\V c_1, \V c_2$ as contact points.

\begin{equation}
\V c_i
	 =\begin{bmatrix}
	1 & 0 & 0 \\
	  0 & 1 & 0 
\end{bmatrix} 
	 \V T_\mathcal{P}
	{^\mathcal{P}\tilde{ \V c_i }}
 \label{arm2center}
, \;
^\mathcal{P}\tilde{ \V c_1 }= 
\begin{bmatrix}
	-r\\
	0 \\
        1
\end{bmatrix} 
,
^\mathcal{P}\tilde{ \V c_2} = 
\begin{bmatrix}
	r\\
	0 \\
        1
\end{bmatrix} 
\end{equation}

\subsection{Dual arm manipulation}

\subsubsection{Equilibrium analysis}
Building upon the concepts introduced in Section II, we proceed to derive the energy (Eq. \ref{Eq:W}) and Hessian specific to dual arm manipulation.
We establish a constant diagonal stiffness matrix $\V {K}_c = \text{diag}(k_c,k_c)$ for spring.
Subsequently, the equilibrium condition (Eq. \ref{Eq: W_z}) becomes,
\begin{align}
\partial_z W (\V z, \V u) &=
\begin{bmatrix}
	k_c(2z_x - u_{1x} - u_{2x})  \\
	  k_c(2z_y - u_{1y} - u_{2y})    \\
        \frac{k_c}{2} (\V u_1 - \V u_2) \cdot  (\V c_1(\V z) - \V c_2(\V z))^\bot
\end{bmatrix}  
\label{Eq:W_z}
\end{align}
where $(\V c_1(\V z) - \V c_2(\V z))^\bot$ denotes the normal vector of $(\V c_1(\V z) - \V c_2(\V z))$.
At each control configuration $\V u$, we fix the control to identify the equilibrium state of $\V z^*$. Equilibrium is achieved when $\partial_z W (\V z^*, \V u) = 0$, placing the system within the equilibrium manifold. We observe, $z^*_x = \frac{u_{1x}+u_{2x}}{2}, z^*_y = \frac{u_{1y}+u_{2y}}{2} $,
which indicate the equilibrium position of peg stays at the center of $\V u_1, \V u_2$. In other words, control $\V u$ is maintained at an equidistant and co-linear position relative to $z^*_x, z^*_y$.

Next, for addressing rotation, we proceed to derive the corresponding equation:
\begin{align}
   \tan(z^*_\theta) &= \frac{ u_{2y}-u_{1y}}{u_{2x}-u_{1x}} \nonumber \\
   z^*_\theta &= \arctan(\frac{u_{2y}-u_{1y}}{u_{2x}-u_{1x}}) \pm m\pi, m = 0,1,2,\cdots
\end{align}
where, $m$ is the so-called multiplicity of equilibria \cite{campolo2023basic}.
This implies that with the same control input $\V u$, the corresponding $\V z$ can attain multiple values at equilibrium. Furthermore, from the last entry of Eq. \ref{Eq:W_z}, the normal of $(\V c_1(\V z) - \V c_2(\V z))$ is perpendicular to $(\V u_1 - \V u_2)$, which also indicates the vector $(\V u_1 - \V u_2)$ must be co-linear with $(\V c_1(\V z) - \V c_2(\V z))$.

\begin{figure}[H]  
  \begin{subfigure}{0.24\textwidth}
    \includegraphics[width=\linewidth]{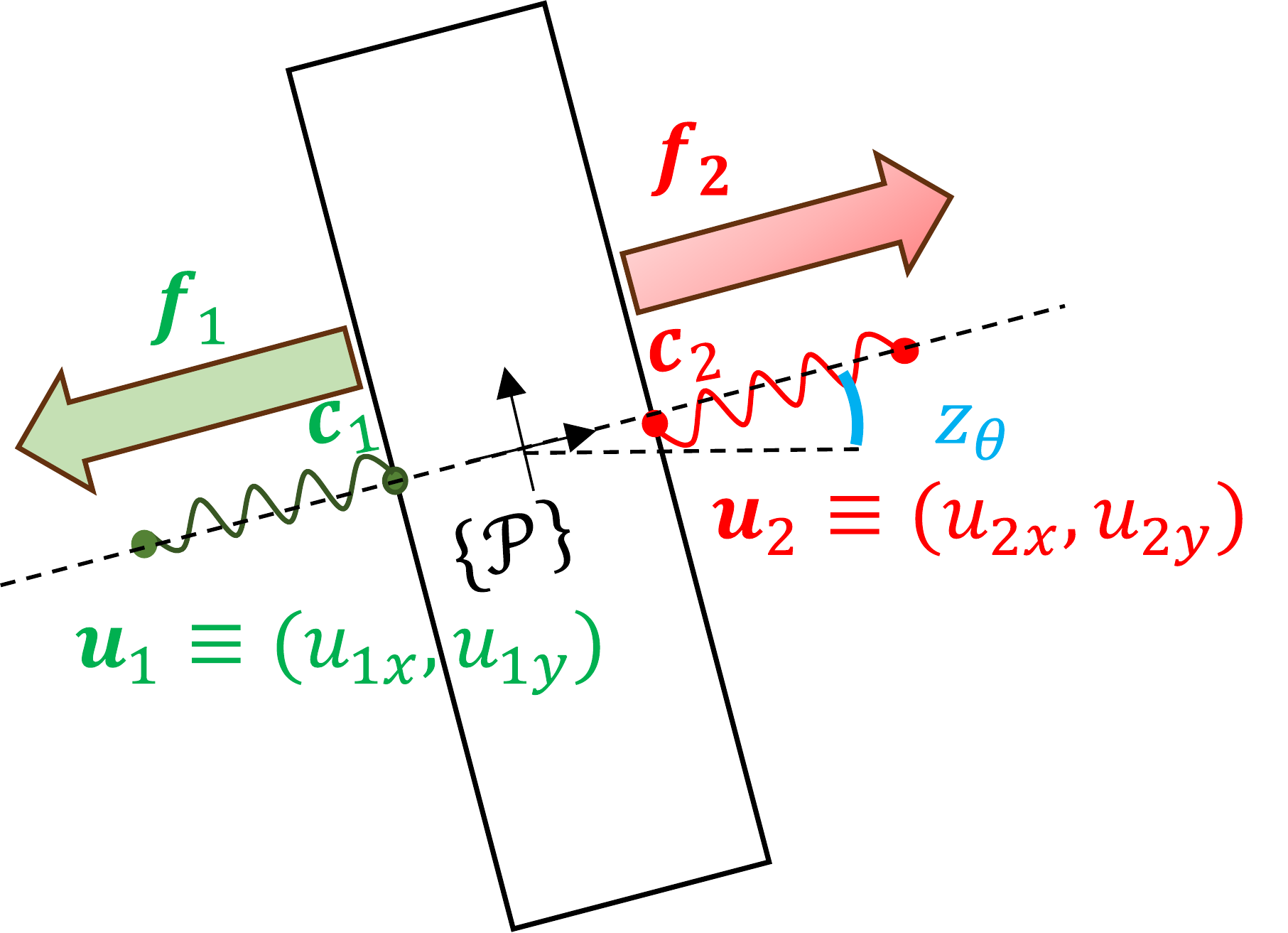}
    \caption[]
    {\small Stable equilibrium.}
    \label{SUBFIG:stable_manipulation}
  \end{subfigure}%
  \hfill  
  \begin{subfigure}{0.24\textwidth}
    \includegraphics[width=\linewidth]{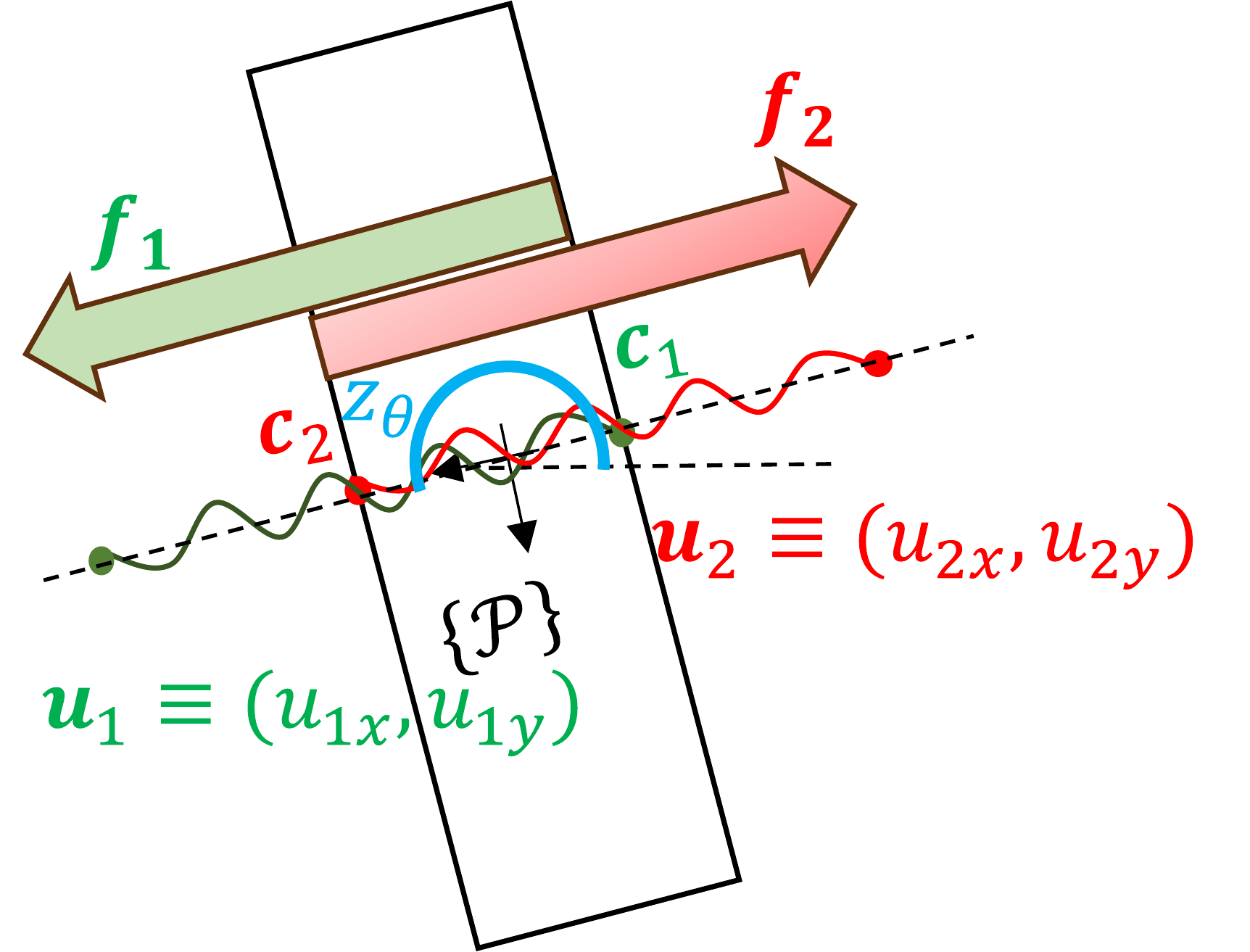}
    \caption[] 
    {\small Unstable equilibrium.}
    \label{SUBFIG:unstable_manipulation}
  \end{subfigure}%
  \vspace{3pt}
\caption{Stable v.s. unstable equilibrium} 
\end{figure}

From a geometric perspective, when $m$ is even (e.g., $m = 0$), the configuration of the system is depicted as Fig. \ref{SUBFIG:stable_manipulation}. In this case, the control inputs spread outward to attain a stable equilibrium, similar to how \textit{a puppeteer pulls strings}.  Conversely, while $m$ is odd (e.g., $m = 1$), the configuration of the system is plotted as Fig. \ref{SUBFIG:unstable_manipulation}. The control inputs cross over each other, strongly squeezing the peg. Hence, at equilibrium $\V u_1 - \V u_2$ is aligned (or anti-aligned) with $\V c_1(\V z) - \V c_2(\V z)$. 
The primary distinction between these two situations lies in whether the two controls $\V u$ intersect, which affects the stability.

\subsubsection{Stability analysis} \label{sec:Stability analysis}


The Hessian of manipulation potential with respect to peg $\V z$ is an important property of the system on the equilibrium manifold. It can be utilized to analyze the stability. 

\begin{align}
\partial^2_{zz} W(\V z, \V u) &= 
\begin{bmatrix}
	2k_c & 0 & 0 \\
	  0 & 2k_c & 0 \\
        0 & 0 & \frac{k_c}{2} (\V u_1 - \V u_2) \cdot  (\V c_1(\V z) - \V c_2(\V z))
\end{bmatrix}  
\label{Eq:Hessian}
\end{align}

The first two terms indicate that two springs connected in parallel along the $x$ and $y$ axis contribute to the total stiffness.
Importantly, the third term emphasizes the alignment requirement, as long as $(\V u_1 - \V u_2)$ must be aligned with $(\V c_1(\V z) - \V c_2(\V z))$, the stability is achieved (Fig. \ref{SUBFIG:stable_manipulation}), while $(\V u_1 - \V u_2)$ is anti-aligned with $(\V c_1(\V z) - \V c_2(\V z))$, the system remains in equilibrium but becomes unstable (Fig. \ref{SUBFIG:unstable_manipulation}).

\LY{Consider the problem of grasping an object with two hands as a control problem, with focus on analysing stability. Even though the inherently stable \textit{mode} for dual arm grasping is shown in Fig. \ref{SUBFIG:stable_manipulation}, this scenario is impractical because it requires the hands to be physically fixed to the object (imagine if your hands were always taped to the object you manipulate in real life). The physically meaningful way to grasp an object is by squeezing it. However, excessive squeezing leads to a mechanically unstable \textit{mode} (as seen in Fig. \ref{SUBFIG:unstable_manipulation}). This is probably why it is difficult to manipulate objects for robots as well.
Our contribution lies in optimising the control policy of a dual arm robot, in this prone to be unstable \textit{mode}, to successfully manipulate the object, while retaining stability.}

Further observation reveals that the magnitude of stability is influenced by several factors: the stiffness of control $k_c$, the size of the peg $\V c_1(\V z) - \V c_2(\V z)$, and the norm of control $\V u_1 - \V u_2$. 
\LY{A higher value of these parameters typically indicates increased system stability. For instance, manipulating a broader object is generally simpler than handling a narrow object, and a higher spring stiffness enhances stability. Spreading the virtual robot control apart usually results in more stability than compressing the object. However, it must be noted that there is no bound on the distance to which the virtual robot control can spread out while keeping the system stable. This is undesirable in practical conditions. For these reasons, the optimal solutions is selected by imposing the friction cone constraint (Eq. \ref{c_fr}) where the robot controls $\V u$ squeeze the object, without cross over each other (Eq. \ref{C_st}).}

\subsection{Stability cost in dual arm manipulation}


To quantify stability in a dual-arm system, we utilize Eq. (\ref{C_st}) and define a reference stiffness matrix $\V K_0 = \text{diag}(k_t,k_t,k_r)$ for peg as a diagonal matrix, where $k_t$ represents translation stiffness while $k_r$ denotes rotation stiffness.
Subsequently, we substitute Eq. \ref{Eq:Hessian} into Eq. \ref{C_st}, which derives a stability cost in a dual-arm peg-in-hole task:
\begin{align}
        \mathcal C_{st} (\V z, \V u) := -\log\left(\frac{2 k_c^2}{k_t^2} \frac{k_c}{ k_r}(\V u_1 - \V u_2) \cdot  (\V c_1(\V z) - \V c_2(\V z))\right)    
    \label{c_st:dual}
\end{align}




To sum up, Eq. \ref{c_st:dual} constrains the relationship between dual arm control $\V u_1, \V u_2$ and the contact point $\V c_1, \V c_2$ on the object, while (\ref{c_fr}) restricts the direction of each control $\V u_i$ with respect to its corresponding contact point $\V c_i$.



\section{Experiment validation} 

\begin{figure}[!h]
    \centering
\includegraphics[width=0.35\textwidth]{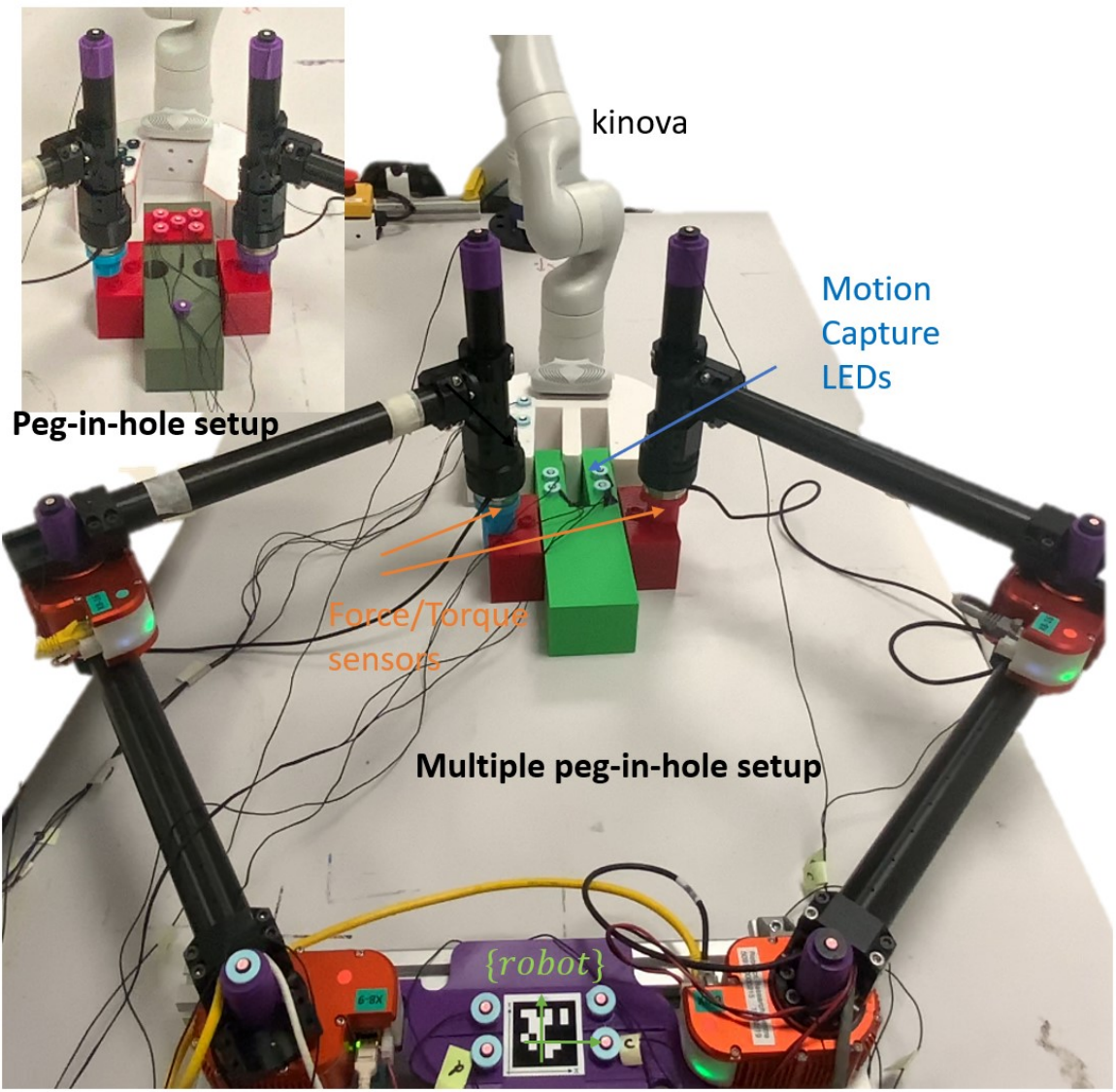}
    \caption{Experimental setup.}
    \label{fig:exp-setup}
\end{figure}

In our experimental setup (Fig. \ref{fig:exp-setup}), we employ DRAKE \cite{drake} as the simulation environment.
\LY{We heuristically choose parameters $d = 1 \; \text{mm}$ (based on the deformation behavior of rubber tape), $l_0 = 25 \; \text{mm}$ (refer to one-third of the size of the peg), $k_t = 10000 \rm{N/m}, k_r = 10 \rm{Nm/rad}$.} 
Moreover, we built a dual arm system consisting of two 2-DOF linkage robots equipped with HEBI joints for the real world experiments to test the optimal trajectories computed by our proposed framework. We equip free handles (in position, but free to rotate, which shift $\V c$ based on width of handle) on the end effector of robot.
To vary the target position/orientation, we utilize a Kinova Gen3 robot with a 3D printed hole fixed. Moreover, motion capture is utilized to record the pose while ATI mini40 is applied to record force. We use 3D printer to prepare a series of hole with different chamfer size.


\subsection{Effect of friction cone in policy optimization}
To study the effectiveness of the friction cone cost (Eq. \ref{c_fr}), we firstly experiment with two distinct types of contact surfaces: plastic-to-plastic and rubber-to-rubber. \LY{Meanwhile, the chamfer is set to $30 \; \text{mm}$ in this comparison to demonstrate the necessity of the friction cone constraint even in a very simple task.}
In the training phase, we implement separate policy for each friction coefficient, characterized (in DRAKE) by $\mu=0.05$ and $0.6$. We symbolize the resulting policies as $\mathrm{DMP}_{0.05}$ and $\mathrm{DMP}_{0.6}$ respectively. 

\begin{figure}[!h]  
\centering
  \begin{subfigure}{0.49\textwidth}
    \includegraphics[width=\linewidth]{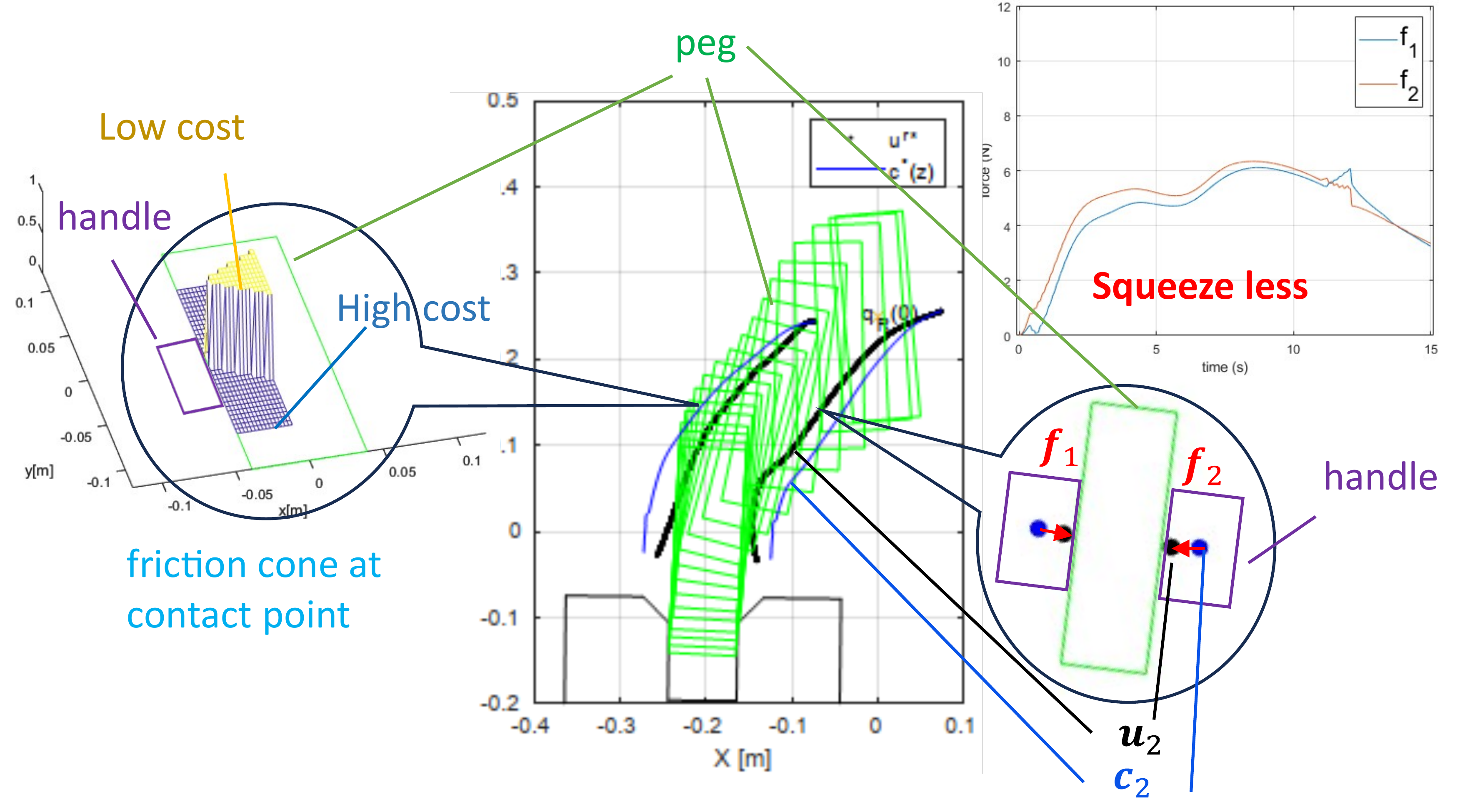}
    \caption[]
    {\small  Left: Cost region for $\mu = 0.6$ with a wide low cost region. Middle: Policy $\mathrm{DMP}_{0.6}$. Bottom right: A single image, dual arm squeeze less. Top right: In the entire manipulation process, high friction coefficient allows less squeeze force for manipulation.}
    \label{SUBFIG:highmu}
  \end{subfigure}%
  \vfill  
  \begin{subfigure}{0.49\textwidth}
    \includegraphics[width=\linewidth]{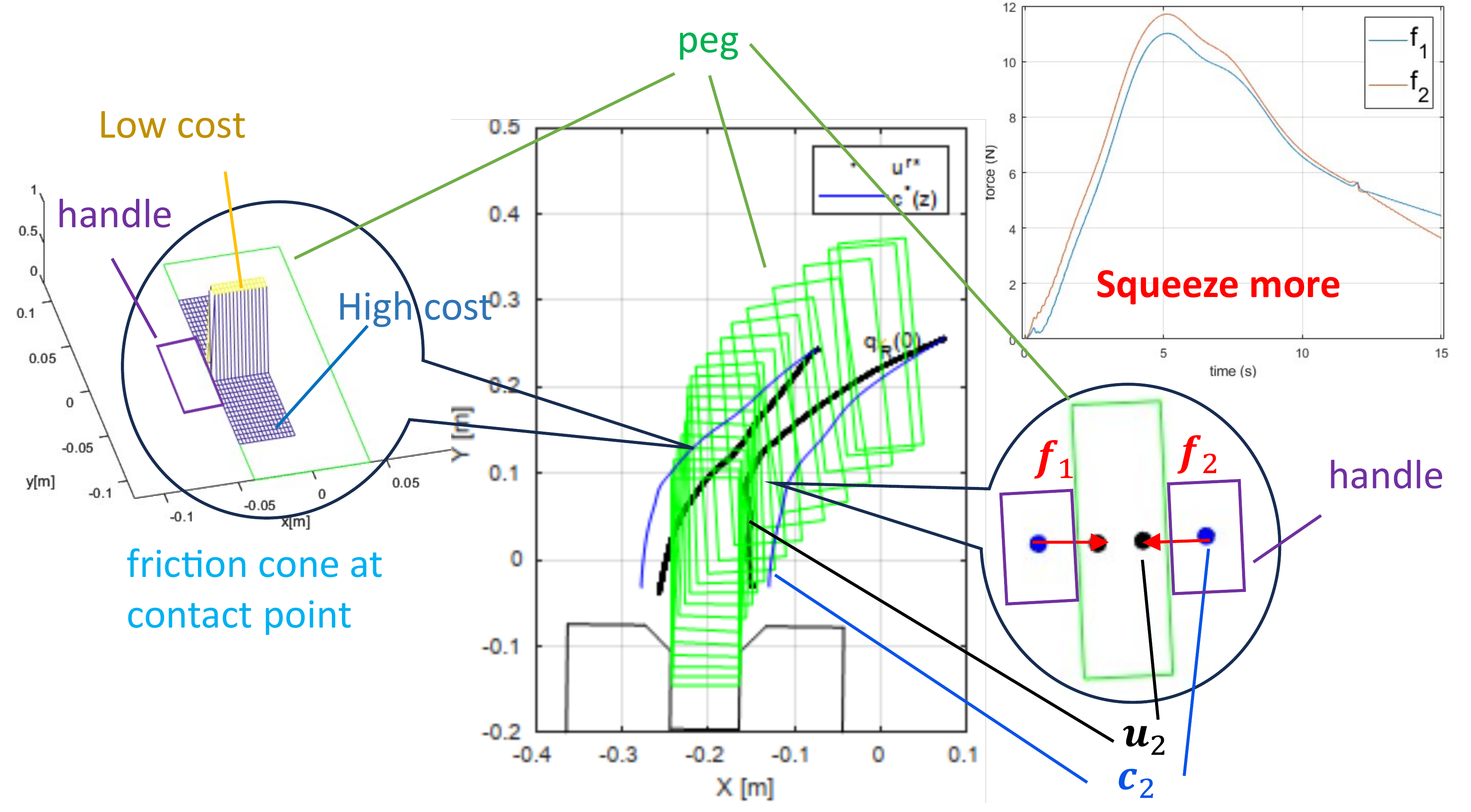}
    \caption[] 
    {\small  Left: Cost region for $\mu = 0.05$ with a narrow low cost region. Middle: Policy $\mathrm{DMP}_{0.05}$. Bottom right: A single image, dual arm squeeze more. Top right: In the entire manipulation process, low friction coefficient requires higher squeeze force for manipulation.}
    \label{SUBFIG:lowmu}
  \end{subfigure}%
\caption{Comparison for DMP with different friction coefficient.} 
\label{fig:compare friction}
\end{figure}

In left side of Fig. \ref{fig:compare friction}, we illustrate the region of low cost $C^{fr}_1$ (the left handle), with above friction coefficient $\mu = 0.6$ (Fig. \ref{fig:compare friction}a) and $0.05$ (Fig. \ref{fig:compare friction}b) respectively.
The green rectangle signifies the peg's geometry. The yellow mesh illustrates the zone where $C^{fr}_1 < 10$, symbolizing regions of lower cost, while the blue mesh represents the high cost region. We notice the small friction coefficient owes a narrow low cost region while the large $\mu$ has a expansive region.

Upon finishing training our framework, the two policies are represented in the middle of Fig. \ref{fig:compare friction}.
When the policy was trained with a higher friction expectation (Fig. \ref{fig:compare friction}(a)), the planning was successful while squeezing less.
In contrast, when expecting lower friction, the optimal policies can be seen to squeeze the peg more (Fig. \ref{fig:compare friction}(b)), i.e., to maximise the resultant tangential friction, so as to avoid slipping. The squeezing force can be observed from the right side of Fig. \ref{fig:compare friction}.

\subsection{$\mathrm{DMP}_{0.05}$ vs $\mathrm{DMP}_{0.6}$ in experiment}
Subsequently, we implement these policies in the real world three separate trials: Policy $\mathrm{DMP}_{0.6}$ on rubber-rubber contact surface, $\mathrm{DMP}_{0.6}$ on plastic-plastic contact surface, $\mathrm{DMP}_{0.05}$ on plastic-plastic contact surface.
During the real world trials, we successfully completed the tasks using the corresponding policies for each specific material combination. 
In the real world experimental validation, while the 3D printed surface finish sufficed for low friction conditions, we pasted a piece of rubber on each hand before the experiment for the high friction condition. To verify the efficacy of the optimal policies, we \textit{cross-tested} their performance, i.e., use $\mathrm{DMP}_{0.6}$ policy in a low friction test. As expected, it was observed the $\mathrm{DMP}_{0.6}$ policy execution failed in the low friction experimental condition, while all the other scenarios succeeded.

\subsection{Effect of stability cost in policy optimization}
\begin{figure}[!h]  
\centering
  \begin{subfigure}{0.24\textwidth}
    \includegraphics[width=0.8\linewidth]{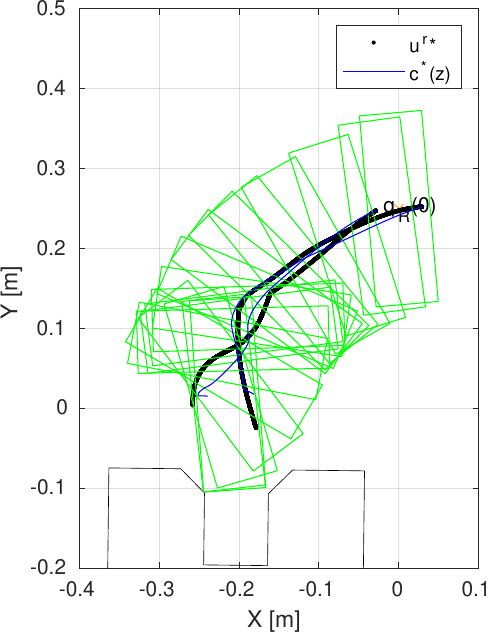}
    \caption{{\small $\mathrm{DMP}_{unst}$: Without stability cost, dual arm cross, causing instability.}}
  \end{subfigure}%
  \hfill  
  \begin{subfigure}{0.24\textwidth}
    \includegraphics[width=0.8\linewidth]{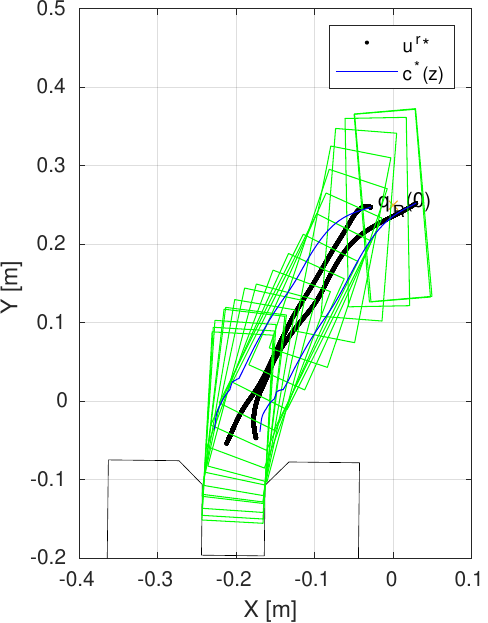}
    \caption{{\small $\mathrm{DMP}_{st}$: With stability cost, dual arm squeeze the peg without crossing, ensuring stability.}}
  \end{subfigure}%
\caption{Effect of stability cost on dual-arm insertion.} \label{fig:unstable}
\end{figure}

\LY{As discussed in in Section \ref{sec:Stability analysis}, unstable manipulation occurs more frequently with thin objects compared to wider ones. Therefore, we define a thin handle to underscore the significance of stability cost (Eq. \ref{c_st:dual}).}
Thus, we computed the optimal policy for conditions $\mathrm{DMP}_{unst}$  (in Fig. \ref{fig:unstable}(a)), a clearly unstable policy, and $\mathrm{DMP}_{st}$ (in Fig. \ref{fig:unstable}(b)), 

The comparative analysis revealed that, the trajectory of the control policy exhibited is notably unstable in the absence of the stability term. Specifically, the trajectories of the control policies tended to over-cross dual arms and result in rotational motion. This observation aligns with the theoretical analysis outlined in Eq. \ref{c_st:dual}.

\subsection{Enhancement by robust training across tasks}

The camera-induced uncertainty is modeled via translation and orientation uncertainties, where standard deviations (STD) $\sigma_x = \sigma_y = 0.34 \, \mathrm{mm}, \sigma_\theta = 3.63^\circ$ (obtained using a simple Apriltag detection test using a calibrated camera). Typically these numbers tally with errors found in literature \cite{morar2020comprehensive}. This setup reflects the real world uncertainty, where the relationship between the actual and estimated poses of the hole follows this distribution, as detailed in Section \ref{sec:DMPBBO}. Therefore, We apply this STD to Gaussian distribution on the hole frame. 
In this task, the optimal policy from our method is termed as robust DMP since our focus is to ensure robustness in an uncertain world. 
Conversely, without our method, the robot would assume the target is accurate. The optimal trajectory from that assumption is called base DMP.
We select $R = 15, \, S = 20$ applying same random seed for all the scenarios in both simulation and real world, while Kinova is controlled by motion capture to achieve same variation as simulation.
We apply our framework to both dual-arm single peg-in-hole and dual-arm multiple peg-in-hole tasks, evaluating its performance with varying chamfer sizes in both simulation and real world experiments. The chamfer-to-peg diameter ratio follows the reference values reported in the literature \cite{hoffman2021control}. The results in simulation and real world are shown in Table \ref{table_result}. Key observations are as follows:

\begin{table}[!t]
\renewcommand{\arraystretch}{1.1}
\caption{Success/Failure: Base DMP (classical method) vs. Robust DMP (proposed).\label{table_result}}
\centering
\resizebox{\linewidth}{20mm}{
\begin{tabular}{ccccc}
\toprule
\textbf{Task}	& \textbf{Environment}	& \textbf{Chamfer-to-peg ratio}	& \textbf{Base DMP } & \textbf{Robust DMP} \\
\midrule
\multirow{4}{*}{Peg-in-hole} & \multirow{3}{*}{Simulation} & 0.385 & 12/20 & 17/20   \\ \cmidrule(lr){3-5}
         &   & 0.256 & 9/20 & 17/20 \\ \cmidrule(lr){3-5}
         &   & 0.128 & 6/20 & 17/20   \\ \cmidrule(lr){2-5}
& Real world & 0.128 & 2/20 & 18/20 \\ \midrule
\multirow{3}{*}{Multiple peg-in-hole} & \multirow{2}{*}{Simulation} & 
             0.192 & 8/20 & 18/20   \\ \cmidrule(lr){3-5}
         &   & 0.128 & 5/20 & 16/20 \\ \cmidrule(lr){2-5}
& \multirow{2}{*}{Real world} & 0.192 & 7/20 & 17/20 \\ \cmidrule(lr){3-5}
& & 0.128 & 1/20 & 16/20 \\ 
\bottomrule
\end{tabular}}
\end{table}

\begin{figure}[!h]  
  \begin{subfigure}{0.24\textwidth}
    \includegraphics[width=\linewidth]{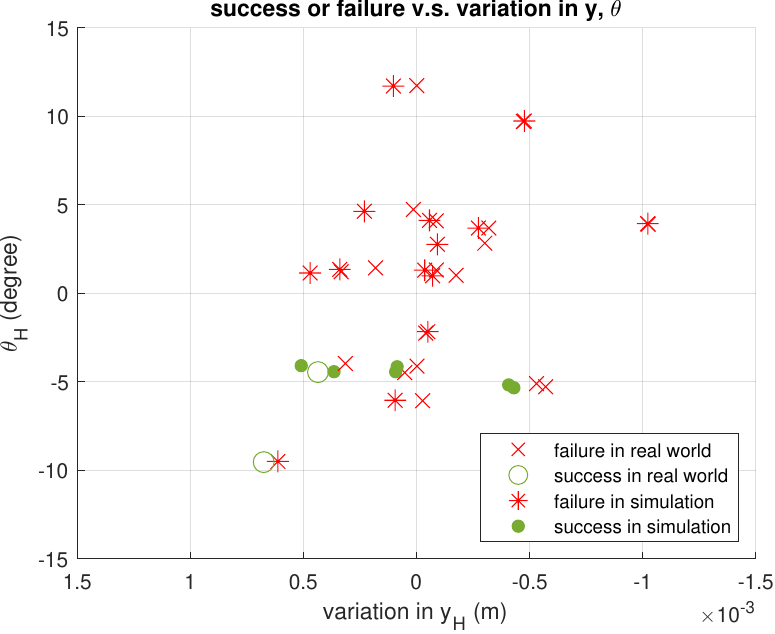}
    \caption[]
    {\small Base DMP across 20 trials.}
  \end{subfigure}%
  \hfill  
  \begin{subfigure}{0.24\textwidth}
    \includegraphics[width=\linewidth]{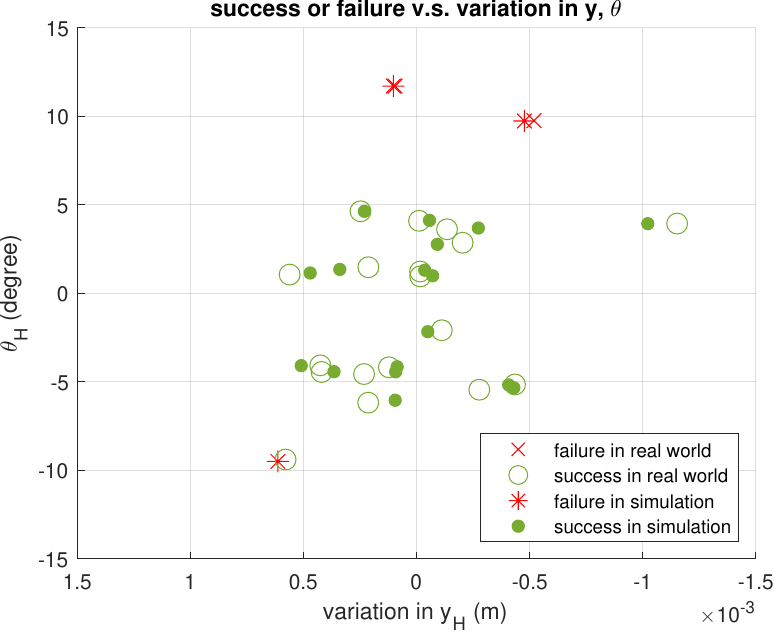}
    \caption[]
    {\small Robust DMP across 20 trials.}
  \end{subfigure}%
  \vspace{3pt}
\caption{Base DMP v.s. robust DMP (dual arm peg-in-hole, chamfer ratio 0.128).} \label{fig:SandF}
\end{figure}

\begin{itemize}
    \item Effect of chamfer:
    As chamfer-to-peg ratio decreases, the success rate of base DMP drops significantly, whereas our robust DMP maintains high performance. This trend is even more pronounced in the multiple peg-in-hole task, where the gap between the two methods widens at smaller chamfer ratios. These results highlight the classical method's susceptibility to failure in small chamfer scenario due to its reliance on accurate estimation, while the robust DMP effectively adapts to different hole poses.

    \item Sim2real analysis: 
    Base DMP suffers a notable performance drop when transitioning from simulation to real world, reflecting its vulnerability to perception errors and uncertainties. In contrast, our robust DMP retains consistently high success rates across this sim2real gap.
    
    \item Peg-in-hole v.s. multiple peg-in-hole: 
    At the same chamfer ratio, multiple peg-in-hole tasks are more challenging than single peg-in-hole tasks. However, a larger chamfer can partially mitigate this difficulty.
    Despite this, the performance gap between the robust and base DMP remains substantial across both tasks, further underscoring the robust DMP’s superior ability to handle uncertainty.

    \item Overall performance and robustness:
    The results show that the robust DMP consistently outperforms the base DMP across all tested conditions, achieving higher success rates despite variations in chamfer size, task complexity, and real world uncertainties. While failures occur under extreme conditions, our method maintains a clear success boundary. Fig. \ref{fig:SandF} illustrates this in one case (peg-in-hole, chamfer ratio 0.128), exemplifying how the robust DMP extends the feasible range of successful execution compared to the base DMP.

\end{itemize}


\vspace{-2mm}

\section{Conclusion}

This study introduces an effective approach that integrates DMP with BBO to manage multi-contact manipulation tasks. By incorporating friction cone, squeeze and stability terms into a robust training methodology, this research establishes a versatile framework capable of addressing a wide array of contact surfaces and accommodating positional inaccuracies. 
Furthermore, the stability of the dual arm peg-in-hole is analytically proven and tested in simulation. 
The results from both simulation and real world experiments show that our robust framework significantly improves insertion success rates over classical DMP. It maintains high performance with decreasing chamfer-to-peg ratio and effectively handles uncertainties that cause failures in traditional methods. The multiple peg-in-hole task further validates its robustness, consistently outperforming the base DMP under increased complexity.

In the future, we will aim to encompass the dual arm grasping phase for multi-manipulator systems, meanwhile, we will involve conducting experiments in 3D space to further validate our framework.

\printbibliography

\end{document}